\documentclass[11pt,a4paper]{article}

\usepackage[nohyperref]{naaclhlt2019}

\usepackage{xcolor}
\definecolor{darkblue}{rgb}{0, 0, 0.5}

\usepackage{url}

\usepackage{breakurl}
\usepackage[breaklinks, colorlinks=true,citecolor=darkblue, linkcolor=darkblue, urlcolor=darkblue]{hyperref}

\usepackage{times}
\usepackage{latexsym}
\usepackage{amsmath}
\usepackage{amssymb}
\usepackage{url}

\aclfinalcopy % Uncomment this line for the final submission

\newcommand{\good}[1]{{\bf #1}}
\newcommand{\bad}[1]{{\textcolor{red}{#1}}}

\title{Alternative Weighting Schemes for ELMo Embeddings}

\author{Nils Reimers \and Iryna Gurevych \\
Ubiquitous Knowledge Processing Lab (UKP-TUDA)\\
Department of Computer Science, Technische Universit\"at Darmstadt\\
\url{www.ukp.tu-darmstadt.de}}

\date{}

\begin{document}
\maketitle
\begin{abstract}
ELMo embeddings \cite{Peters_2018} had a huge impact on the NLP community and may recent publications use these embeddings to boost the performance for
downstream NLP tasks. However, integration of ELMo embeddings in existent NLP architectures is not straightforward. In contrast to traditional word embeddings, like GloVe or word2vec
embeddings, the bi-directional language model of ELMo produces three 1024 dimensional vectors per token in a sentence. Peters et al.\ proposed to learn a task-specific weighting of these three vectors for downstream tasks. However, this proposed weighting scheme is not feasible for certain tasks, and, as we will show, it does not necessarily yield optimal performance. We evaluate different methods that combine the three vectors from the language model in order to achieve the best possible performance in downstream NLP tasks. We notice that the third layer of the published language model often decreases the performance. By learning a weighted average of only the first two layers, we are able to improve the performance for many datasets. Due to the reduced complexity of the language model, we have a training speed-up of 19-44\% for the downstream task.\footnote{Code: \href{https://github.com/UKPLab/elmo-bilstm-cnn-crf}{https://github.com/UKPLab/elmo-bilstm-cnn-crf}} 
\end{abstract}

\section{Introduction}

\newcite{Peters_2018} presented in their work \textit{Deep Contextualized Word Representations} (often referred to as ELMo embeddings) a method that uses a bidirectional language model (biLM) to derive word representations which are based on the complete context of a sentence. They demonstrated that these ELMo embeddings can substantially increase the performance for various NLP tasks. This new type of word representations had a big impact on the NLP community and many new architectures, for example, many from EMNLP 2018, report a better performance when using ELMo embeddings.

Traditional word embeddings, like word2vec or GloVe, mapped each token in a sentence to a single dense vector. In contrast to that, the published ELMo implementation computes three layers of a language model: The first layer is a CNN that computes a non-contextualized word representation based on the characters of a word, followed by two bidirectional LSTM layers that take the context of the sentence into account.

The output of the three layers is integrated into task-specific neural architectures. However, the integration of ELMo into neural architectures is not straightforward. For example, Peters et al.\ describe two methods for the integration: Either the output of the last layer is used for downstream tasks, or a task-specific weighting of the three layer outputs is learned:
$$\text{ELMo}_{\text{weighted\_average}} = \gamma \sum_{j=0}^{2}s_j h_j$$ 
with $s \in \mathbb{R}^3$ softmax-normalized weights, $h_j$ the output of the $j$-th layer of the biLM and  $\gamma$ a scalar that is used to scale the entire ELMo vector.

Learning this weighted average is not always easy, as it can require substantial changes in existent network architectures and some deep learning frameworks (for example Keras) lack the possibility to easily implement such a learned weighted average. Further, for unsupervised tasks, such a weighted average cannot be learned.

Hence, several authors used simplified ways to integrate ELMo embeddings in their architectures. Some use the output of the last ELMo layer, some concatenate all three vectors \cite{elmo_concat1,elmo_concat2}, and others compute a (fixed) average of the three layers.

It is unclear what the impact of these different weighting schemes is for downstream tasks. Is the (rather complicated) learned weighted average proposed by Peters et al.\ needed to achieve optimal performance? Will a simpler method, like computing a fixed average, decrease the performance?

In this paper, we evaluate different schemes to combine the three ELMo vectors. We analyze the impact of these schemes for downstream NLP tasks. First, we study this for a BiLSTM-CRF architecture which only uses ELMo embeddings as input representation. Next, we study the different weighting schemes for the more complex models included in AllenNLP, which concatenate ELMo embeddings with other input representations like GloVe word embeddings. 

In this paper we show that 1) the weighting scheme can have a significant impact on downstream NLP tasks, 2) that the learned weighted average proposed by Peters et al.\ does not yield  the optimal performance for all datasets, and 3) that the second layer of the biLM yields in many cases a better performance than the third (last) layer.

Surprisingly, using the output of the second layer of the biLM model yields a better performance than using the third (last) layer in many downstream NLP tasks. Using this insight, we present a weighting scheme that learns a weighted average of the first two layers of the biLM. This scheme outperforms the originally proposed weighting scheme by Peters et al.\ for several datasets. Further, it is computationally faster than the original method. For downstream tasks,  we saw a training speed-up of 19-44\%.

\section{Related Work} 
To our knowledge, only \newcite{Peters_2018} evaluated different weighting schemes. They evaluated to use either the output of the last layer or to learn a task-specific weighted average of all three layer outputs. They compare these two options in their paper and show a slight advantage for learning a weighted average. However, the evaluation is in our opinion insufficient. First, they evaluate both options on the development set, so it remains unclear if there are changes for unseen data (test set). Further, they evaluate it only with a single random seed. As shown in \cite{Reimers_2018_single_performance_scores}, the performance of a neural network can change significantly with a different random seed. For example,  we observe test score differences of up to 1.5 percentage points when the same model is trained with a different random seed with the AllenNLP model for the Stanford Sentiment Treebank (SST-5). The differences Peters et al.\ report between using the last layer and learning a task-specific weighting are rather small (0.4 - 0.7 percentage points). It is not clear if these differences are due to the effect of different random seeds or due to the weighting scheme.

\section{Alternative Weighting Schemes}
The published bidirectional language model (biLM) produces three 1024 dimensional vectors for each token in a sentence. In this paper we systematically study the following methods to combine the three vectors returned by the biLM:

\noindent \textbf{Individual Layers:} Only a single layer is used for the downstream task.
 
\noindent \textbf{Concatenation:} All three vectors are concatenated. 
 
\noindent \textbf{Fixed Average:} We compute an average of all three vectors. 
 
\noindent \textbf{Learned Weighted Average:} We learn a task-specific weighted average (ELMo$_\text{weighted\_average}$).

\noindent \textbf{Learned Weighted Average of the 1st and 2nd Layer:} We learn a task-specific weighted average of the first two layers.

\begin{table*}[t]
\begin{tabular}{|l|c|c|c|c|c|c|c|}
\hline
\textbf{Dataset} & \textbf{1. Layer} & \textbf{2. Layer} & \textbf{3. Layer} & \textbf{Concat} & \textbf{F.-Avg.} & \textbf{W.-Avg.}  & \textbf{W.-Avg. 1 \& 2} \\ \hline
Arguments & \bad{56.89} & 58.84 & 57.48 & 57.82 & \good{59.03}  & 58.75  & 59.30 \\ \hline
ACE Entities & \bad{84.28} & \good{86.07} & \bad{85.05} & 85.94 & 86.05 & 85.86  & 86.01 \\ \hline
ACE Events  & \bad{64.88} & 67.66 & \bad{65.54} & \good{67.94} & 67.74 & 67.25 &  67.69 \\ \hline
POS  & \bad{90.25} & \good{92.33} & \bad{90.46} & 92.21 & \bad{91.75} & \bad{91.94} &  \bad{91.48} \\ \hline
Chunking  & \bad{95.07} & \good{96.55} & \bad{96.14} &  \bad{96.38} & \bad{96.34} & 96.52 & 96.51 \\ \hline
NER  & \bad{90.50} & 91.78 & \bad{91.53} & 91.67 & \good{91.81} & 91.73 &  91.76 \\ \hline
GENIA NER  & 72.77 & \bad{72.09} & \bad{70.59} &  72.69 & 72.41 & \good{73.08} &  73.26 \\ \hline
WNUT16 NER  & \bad{44.83} & 49.88 & \bad{48.58} &  49.53 & 48.72 & 49.56 & \good{49.90} \\ \hline \hline 

 NER & \bad{91.17} & 92.24 & 92.28 & 92.09 & \good{92.46} & 92.34 & 92.14 \\ \hline 
 GENIA NER  & 74.00 & 73.94 & 73.76 & 74.09 & 73.68 & 74.08 & \good{74.16} \\ \hline
 SST  & \bad{51.76} & 53.41 & 52.91 & 53.86 & 53.19 & 53.33 & \good{54.17} \\ \hline 
 Parsing & \bad{92.51} & \bad{94.10} & \bad{93.84} & \bad{94.09} & \bad{93.94} & 94.37 & \good{94.39} \\ \hline
 SNLI & 88.03 & 88.11 & \bad{87.84} & 88.48 & 88.43 & \good{88.50} & 88.32 \\ \hline
\end{tabular}
\caption{Upper half: Average scores of 10 training runs with different random seeds using the BiLSTM-CRF architecture of \newcite{Reimers_2017_EMNLP}. ELMo is used as the only input representation. Lower half: Average scores of 10 training runs using AllenNLP. ELMo is used in combination with other input representations. Bold entries mark the best performance per row. Red entries indicate statistically significantly worse entries with $p<0.01$. \textit{F.-Avg.}: unweighted average of all three layers, \textit{W.-Avg.}: Learned weighted average of the three layers, as proposed by Peters et al., \textit{W.-Avg. 1 \& 2:} Learned weighted average of the first and second layer of the biLM.}
\label{table_weighting_schemes}
\end{table*}

\section{Evaluation of Weighting Schemes}
We test the different weighting schemes with two experiments. For the first experiment, we evaluate a neural network that solely uses ELMo embeddings as a representation of the input. This experiment shows how suitable the schemes are when no other features are used. In the second experiment, we evaluate the schemes with the more advanced, state-of-the-art architectures from AllenNLP\footnote{\url{https://github.com/allenai/allennlp}}. These models often concatenate the ELMo embeddings with other input representations. For example, the NER model from AllenNLP concatenates the ELMo embedding with GloVe embeddings and with a task-specific character-based word representation (similar to \newcite{Ma2016}). We expect that the results in the second experiment vary from the first experiment. If a particular weighting scheme lacks specific information, the network might still retrieve it from the other input representations. 

For the first experiment, we use a BiLSTM-CRF architecture for sequence tagging \cite{Huang2015}. We use ELMo embeddings instead of word embeddings. Two bidirectional LSTM layers (with 100 recurrent units each) are followed by a conditional random field (CRF) to produce the most likely tag sequence. The network was trained using Adam optimizer \cite{Adam} and a variational dropout \cite{Gal2015} of 0.5 was added to recurrent and output units.

We trained this architecture for the following datasets: \textbf{Arguments:} Argument component detection (major claim, claim, premise) in 402 persuasive
essays \cite{Stab_2017}. Development and test set were 80 randomly selected essays each. \textbf{ACE Entities/Events:} ACE 2005 dataset \cite{ace_2005} consists of 599 annotated documents from six different domains (newswire, broadcast news, broadcast conversations, blogs, forums, and speeches). We train the architecture to either detect events or to detect entities in these documents. We used 90 randomly selected documents each for the development and test set.  \textbf{POS:} We use the part-of-speech tags from Universal Dependencies v.\ 1.3 for English with the provided data splits. We reduced the training set to the first 500 sentences to increase the difficulty for the network. The development and test set were kept unchanged.  \textbf{Chunking:} CoNLL 2000 shared task dataset on chunking. \textbf{NER:} CoNLL 2003 shared task on named entity recognition. \textbf{GENIA NER:} The Bio-Entity Recognition Task at JNLPBA \cite{genia_ner} annotated Medline abstracts with information on bio-entities (like protein or DNA-names). The dataset consists of 2000 abstracts for training (we used 400 of those as development set) and the test set contains 404 abstracts. \textbf{WNUT16:} WNUT16 was a shared task on Named Entity Recognition over Twitter \cite{WNUT16}. Training data are 2,394 annotated tweets, development data are 1,000 tweets, and test data are 3,856 tweets.

For the second experiment, we use the existent AllenNLP models that reproduce the experiments of Peters et al. We use the CoNLL 2003 NER model, the Stanford Sentiment Treebank (SST-5) model, the constituency parsing model for the Penn TreeBank, and the Stanford Natural Language Inference Corpus (SNLI) model. The $F_1$-score is computed for the NER tasks and parsing; accuracy is computed for the SST-task and the SNLI-task.

Not all experiments from the paper of Peters et al.\ are reproducible with AllenNLP. AllenNLP currently has no model for the SQuAD task. For the Coref-task, the AllenNLP configuration is missing some features and does not use ELMo embeddings. For the SRL-task, AllenNLP uses a different metric that is not comparable to the official metric.

For both experiments, we use the pre-trained ELMo 5.5B model, which was trained on a dataset of 5.5 billion tokens. We trained each setup with ten different random seed and report average test scores.

\subsection*{Results}

The results of the BiLSTM-CRF model, that uses only the ELMo embeddings as input representations, are shown in the upper part of Table \ref{table_weighting_schemes}.

We observe that the output of the first layer yields in most cases the worst performance. This was expected, as the first layer is a CNN that computes a character-based word representation. It does not take the context of a word into account. 

In our experiment, we don't observe a difference between the computation of an unweighted average and of learning a task-specific weighted average. For four datasets, the unweighted average yielded better performance, while for the other four other datasets, the learned weighted average yielded better performance. However, the differences are insignificant.

To our surprise, using the second layer of the biLM yields in most cases a better performance than using the third (last) layer of the biLM. For 7 out of 8 datasets it outperforms even the learned weighted average method proposed by Peters et al. Only for the GENIA dataset achieved the learned weighted average a significantly better performance than using the output of the second layer. However, for this dataset, it appears that context information is less critical, as the system achieves a high performance by using only the characters of a word (1. Layer).

The results for the second experiment, that uses AllenNLP and ELMo embeddings in combination with other input representations, are presented in the lower part of Table \ref{table_weighting_schemes}.

In contrast to our first experiment, we notice much smaller differences between different weighting schemes. In most cases, the differences are not statistically significant. When a network solely depends on ELMo embeddings as input, all relevant information to solve the task must be included. However, if ELMo is combined with other input representations, like GloVe embeddings, the dependency on the ELMo embeddings decreases. If ELMo does not capture critical word properties, the network can still fall back on one of the other input representations.

We notice that computing a weighted average of the first two layers yields the highest performance in 4 out of 5 datasets. It appears that computing the third layer of the biLM does not necessarily add value for downstream tasks. Removing it increases training as well as inference speed, as computing the bidirectional LSTM of the third layer is rather slow. By not computing the third layer of the biLM, we observe a training speed-up of 38\% for the NER model, 44\% for the SST model, 19\% for the parsing model, and  27\% for the SNLI model. The training of the SNLI model requires on a Tesla P100 GPU about one day.

\section{Conclusion}

As noted by \newcite{Peter_2018_b}, the lower layers of the biLM specialize in local syntactic relationships, allowing the higher layers to model longer range relationships. Knowing beforehand which of these properties are relevant for an NLP task is difficult. In our experiments, we observed significant performance differences in downstream tasks for the three biLM layers. The last, most abstract layer, often yielded mediocre results when it was used as the only input representation. For most datasets, the output of the second layer appears to be the most relevant representation. It offers the best trade-off between local syntactic features and more abstract, long-range relationships.

As it is not known in advance which layer produces the best input representation, learning a task-specific weighted average of the three layers appears advisable. However, for several datasets, it appears that the output of the third layer does not add value as it is a too abstract representation for the NLP tasks. The learned weighted average method presented by Peters et al.\ regularizes the three (softmax-normalized) weights $s_j$. As a consequence, a zero or small $s_j$ value is not possible, and all three vectors are used even if one vector (e.g.\ the third layer output) decreases the performance. This explains why removing the (sometimes harmful) third layer can improve the model performance. Further, by removing the last layer, we observe a significant training speed-up. For the models included in AllenNLP, we observed a training speed-up of 19-44\%, while improving the test performance in 3 out of 5 datasets. This speed-up can be crucial for cases that require fast training of inference speeds.

The weighting scheme appears especially important when these vectors are used as the only input representation for the task. In that case, we advise testing different weighting schemes. If ELMo is used in conjunction with other input representations, the weighting scheme was less critical.

\section*{Acknowledgements}
This work was supported by the German Research Foundation through the German-Israeli Project Cooperation (DIP, grant DA 1600/1-1 and grant GU 798/17-1).  We gratefully acknowledge the support of NVIDIA Corporation with the donation of the Titan X(p) Pascal GPU used for this research.

%\newpage

\bibliography{naaclhlt2019}
\bibliographystyle{acl_natbib}

\appendix

\end{document}